\documentclass[smallcondensed, natbib]{svjour3}
\usepackage[latin9]{inputenc}
\usepackage{color}
\usepackage{array}
\usepackage{wrapfig}
\usepackage{textcomp}
\usepackage{amsmath}
\usepackage{amssymb}
\usepackage{graphicx}
\usepackage[unicode=true,
 bookmarks=false,
 breaklinks=false,pdfborder={0 0 0},backref=section,colorlinks=true]
 {hyperref}

\usepackage{array}
\usepackage{lineno}
\modulolinenumbers[5]

\usepackage{todonotes}
\usepackage{listings}

\usepackage[export]{adjustbox}
\usepackage{xcolor}
\usepackage{booktabs}

\usepackage{listings}

\def\makeheadbox{\relax}

\begin{document}

\title{Kilombo: a Kilobot simulator to enable effective research in swarm
robotics}

\author{Fredrik Jansson\thanks{F Jansson and M Hartley contributed equally to this work.} \and Matthew Hartley \and Martin Hinsch 
\and Ivica Slavkov \and Noemí Carranza
\and Tjelvar S. G. Olsson \and Roland M. Dries \and Johanna H. Grönqvist
\and Athanasius F. M. Marée \and James Sharpe \and Jaap A. Kaandorp
\and Verônica A. Grieneisen}

\institute{F Jansson \and R Dries \and J Kandorp \at Computational Science Lab, University of Amsterdam, The Netherlands\\\email{fjansson@abo.fi}
\and
M Hartley \and M Hinsch \and T Olsson \and A Marée \and V Grieneisen \at John Innes Centre, Norwich Research Park, Norwich NR4 7UH, United Kingdom \\\email{Matthew.Hartley@jic.ac.uk}, \email{Veronica.Grieneisen@jic.ac.uk}
\and 
I Slavkov \and N Carranza \and J Sharpe \at Centre for Genomic Regulation (CRG), The Barcelona Institute of
Science and Technology, Dr.~Aiguader 88, 08003 Barcelona, Spain
\and 
I Slavkov \and N Carranza \and J Sharpe \at Universitat Pompeu Fabra (UPF), Barcelona, Spain
\and
J Sharpe \at Institució Catalana de Recerca i Estudis Avançats (ICREA), Passeig
Lluís Companys 23, 08010 Barcelona, Spain
\and 
R Dries \at Department of Bionanoscience, Delft University of Technology, The
Netherlands 
\and 
J Grönqvist \at Department of Physics, Åbo Akademi University, Turku, Finland}

\authorrunning{Jansson et al.}
\maketitle

\begin{abstract}
The Kilobot is a widely used platform for investigation of swarm robotics.
Physical Kilobots are slow moving and require frequent recalibration
and charging, which significantly slows down the development cycle.
Simulators can speed up the process of testing, exploring and hypothesis
generation, but usually require time consuming and error-prone translation
of code between simulator and robot. Moreover, code of different nature
often obfuscates direct comparison, as well as determination of the
cause of deviation, between simulator and actual robot swarm behaviour.
To tackle these issues we have developed a C-based simulator that
allows those working with Kilobots to use the same programme code
in both the simulator and the physical robots. Use of our simulator,
coined Kilombo, significantly simplifies and speeds up development,
given that a simulation of 1000 robots can be run at a speed 100 times
faster than real time on a desktop computer, making high-throughput
pre-screening possible of potential algorithms that could lead to
desired emergent behaviour. We argue that this strategy, here specifically
developed for Kilobots, is of general importance for effective robot
swarm research. The source code is freely available under the MIT
license. 
\end{abstract}



\section{Introduction}

\label{introduction}


Research employing swarms of robots has in the recent years become
increasingly important, as it allows one to investigate how complex
emergent behaviour can be generated by many interacting agents that
self-organize in a non-hierarchical and distributed manner. For example,
swarms of physical robots were able to challenge current perceptions
concerning how robustness in problem solving can come forth due to
local interactions \citep{Valentini.tnac15,Ferrante.pcb15}. Testing
algorithms for self-organization on actual physical robots constitutes
the ultimate proof-of-principle of the thought-out concepts, mechanisms
and hypotheses on which those algorithms are based. However, to do
so in an efficient and high-throughput manner, it is of paramount
importance to have tools available in between the drawing board and
the physical robots themselves. 

Here we present one such tool that we have generated, a novel simulator,
coined Kilombo, which allows for linking conceptual ideas to physical
robot swarms. 

\begin{figure}
\centering{}\includegraphics{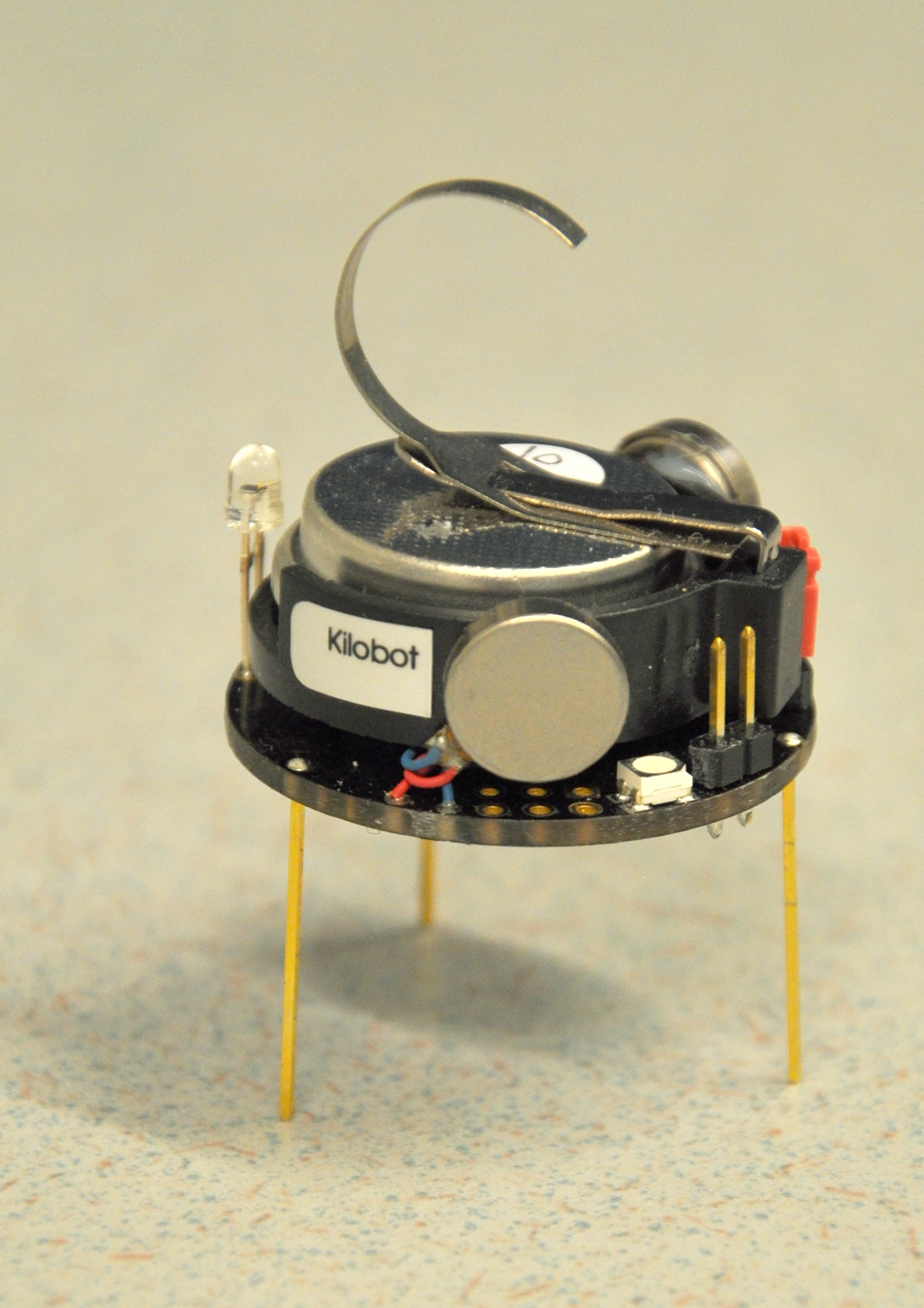} \caption{A Kilobot robot. The robot is supported by three stiff legs, and moves
using a pair of vibration motors.}
\label{fig-kilobot} 
\end{figure}

The Kilobot \citep{Rubenstein2011,Rubenstein2012,Rubenstein2014}
is a low-cost robot developed by the Self-organizing Systems Research
Group at Harvard University, and manufactured by K-team.
They have become the current paradigm system to address self-organization
in large swarms of robots. The robot, shown in Fig.~\ref{fig-kilobot},
is designed for use in large robot swarms: all routine operations
such as programming, switching the robots on or off, and recharging
the battery can be performed at a large scale without handling individual
robots. The capabilities of each individual Kilobot are limited, but
sufficient to implement collective behaviour algorithms, for example
the S-DASH algorithm for shape formation \citep{SDASH2010,DASH2009}.
Since the release of the hardware specifications, Kilobots have been
used in various research projects, e.g.~on collective transport \citep{Rubenstein2013},
control theory \citep{Dorigo-etal2014}, education \citep{zhengwei2014}
and efficient decision making \citep{valentini2015}.

Testing and debugging an algorithm on a collection of Kilobots can
be a challenging task. Not only does it take a considerable amount
of time and effort to set up and program the robot swarm, but running
the swarm algorithm can itself be time consuming. Additionally, in
case of an error or unexpected behaviour it is often difficult to
deduce the reasons for the problem from the behaviour of the robots
alone. Before doing physical experiments it is therefore useful to
test algorithms in a simulator on a PC or a workstation. The simulation
will usually be significantly faster than the actual system and --
in case a problem arises -- allow for detailed inspection and controlled
repetition.

To our knowledge there are at this point two simulators available
for the Kilobot, V-REP and KBsim. V-REP is a comprehensive generic
robot simulation framework \citep{VREP}. It is supported by the Kilobot
designers, who provide a V-REP model of the Kilobot. The recommended
way to interface custom robot code with V-Rep is by implementing scripts
in LUA, although modules in C/C++ are supported as well. The main
part of V-REP is licensed under the GPL, some of the add-on modules
are closed-source however.

KBsim \citep{KBsim} is written in Python and implements a physical
model of the Kilobots. The objective of KBsim is to simulate the underlying
swarm algorithms rather than the actual Kilobot code itself. User
programmes for the KBsim simulator have to be written in Python and
therefore need to be translated afterwards into C in order to be compatible
with and run on the physical Kilobots. Besides, KBsim targets an older
version of the Kilobot library and appears to be currently unmaintained.

Both simulators have in common that simulation speed is suboptimal
due to the overhead imposed by using a scripting language and -- in
the case of V-REP -- the heavyweight physics engine underlying the
simulator. In addition, testing a given algorithm requires two more
or less independent implementations in both cases -- one that is compatible
with the simulator and one that can be compiled to a binary that will
run on the Kilobots. Not only does maintaining two codes increase
the amount of effort required for a given project, but it also presents
an additional source of errors which can be difficult to find. Moreover,
when the research goal is to explore constructive and robust emergent
behaviour, given the diverse sources of noise that multi-agent robotic
systems present, then discrepancies between simulated swarm behaviour
and physical swarm behaviour can be highly informative, but only if
the origin of the discrepancies can be systematically studied and
other potential sources underlying the divergence (such as code differences)
can be eliminated. Therefore, to address fundamental questions in
self-organizing swarm robotics, it is important that the codes used
in the simulator and for the actual robots can be as similar and transparent
as possible.

\begin{figure}
\centering{}\includegraphics[width=1\columnwidth]{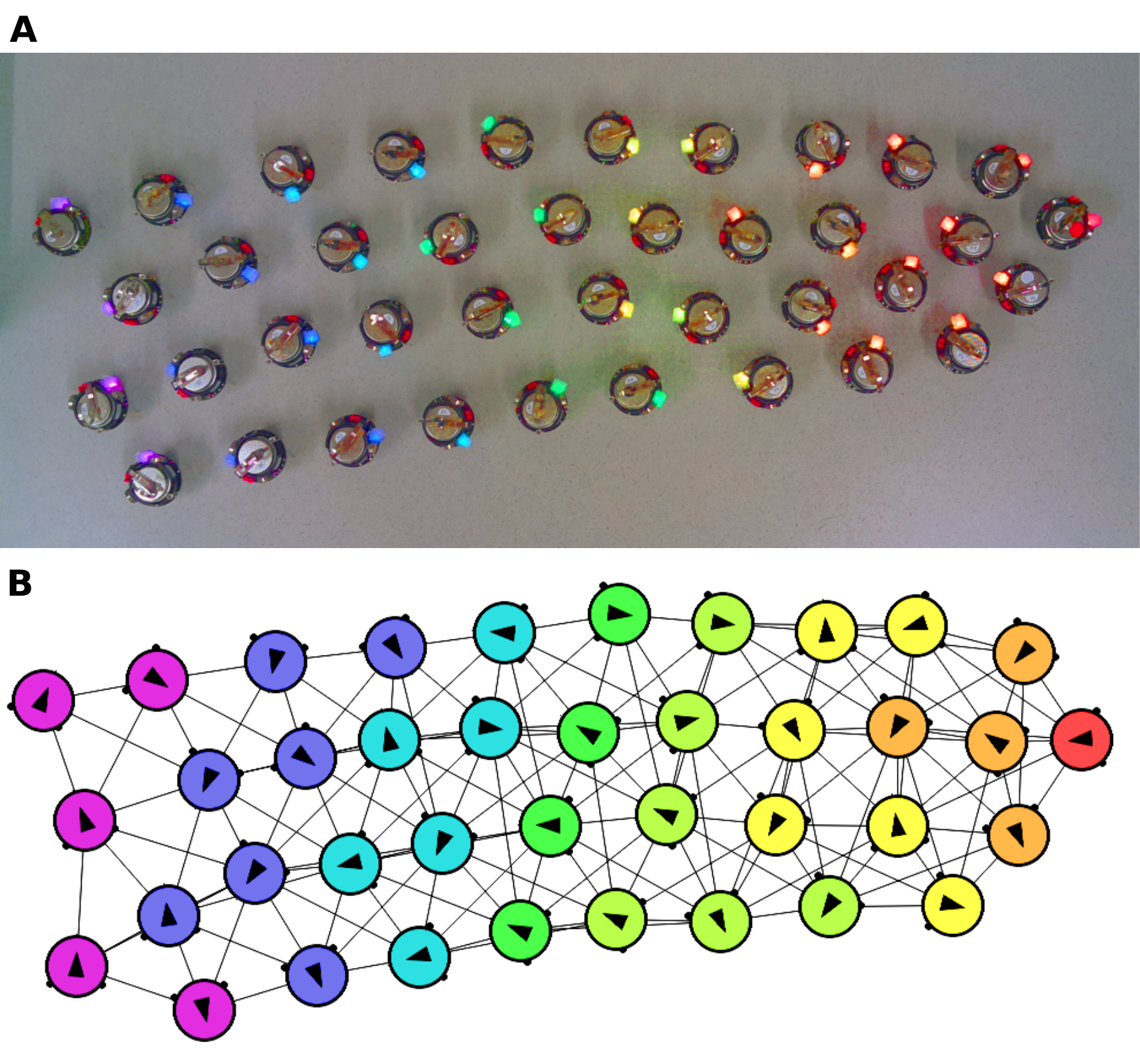}
\caption{Example of a robot swarm propagating a signal that forms a gradient.
\textbf{(A)} The user programme running on real Kilobots; \textbf{(B)}
The same programme running in the simulator.}
\label{fig-simulator} 
\end{figure}

We therefore developed a new simulator, Kilombo, with the aim of a)
being sufficiently accurate, b) being efficient, c) making it possible
for simulator user programmes to run on the physical robots as well,
and d) allowing for already existing Kilobot code to run on the simulator
with as few changes as possible, our overarching goal being to address
a new level of research questions regarding emergent behaviour in
robotics (Fig.~\ref{fig-simulator}).

In this paper we present the design motivation, the general structure
and the usage of the Kilombo simulator. The simulator and its source
code are available on GitHub \citep{sourcecode}, under the MIT license.
The simulator has been tested on various Linux distributions and on
OSX.

The remainder of the paper is organized as follows. Sect.~\ref{design}
describes the design and implementation of the simulator, while Sect.~\ref{portability}
describes how to construct a user programme that can be run both in
the simulator and on a real Kilobot. Comparisons between the simulator
and real robots running example programmes are given in Sect.~\ref{examples}.
These examples allude to the potential research that is unleashed
with such a robot-simulator framework. Performance measurements of
the simulator are shown in Sect.~\ref{efficiency}. A discussion of the
implementation decisions and possible extensions in Sect.~\ref{discussion}
concludes the paper.

\section{Design}

\label{design}

\subsection{Kilobots}

\label{programming-kilobots}

Each Kilobot contains an Atmel ATmega328P microcontroller, which is
programmable in the C language. The robots are equipped with LEDs,
ambient light sensors and short-range infrared communication facilities.
They move on stiff legs with the help of vibration motors.

The user programme running on the robot is compiled from C or C++
with the standard AVR tool chain, based on avr-gcc. The robot hardware
and Kilobot-specific functions are handled using a custom C API (Application
Programming Interface) provided by the Kilobot team, in the form of
a library named kilolib \citep{API}. The functions provided include
controlling the motor speeds, transmitting and receiving infrared
communication, measuring the ambient light level, and setting the
colour of an RGB status LED. When the user C programme is compiled
and linked with kilolib, a programme which can be run on a swarm of
Kilobot robots is generated.

\subsection{Requirements}

We defined three key requirements that our simulator needed to fulfil:
\begin{itemize}
\item Simulations have to capture the essential features of the real Kilobots
-- in particular in scenarios with many interacting robots -- with
sufficient accuracy. Ideally it should be possible to substantially
explore possible behaviours with developed robot code using only the
simulator. Note, however, that it is hard to predict at forehand,
as well as highly algorithm-dependent, to which extent small errors
and variations brought forth by the real-world embedding of real Kilobots
can propagate through the multi-component interactions, thereby both
quantitatively and qualitatively affecting the overall dynamics. We
therefore still consider the performance in the real robot swarm as
a fundamental step in the research, to be contrasted to what has been
observed in the simulator, with larger discrepancies pointing towards
algorithms that are less robust against the complex and multi-modal
noise of `real life' \citep{Jakobi1997}. 
\item Simulations, to be useful, should run significantly faster than the
real system. Faster simulations lead to shorter iteration cycles during
development and debugging, and thus to greater development speed.
In addition, fast simulations open up new types of questions that
the Kilobot system can be applied to.
\item When porting user code from the simulator to the robots (or the other
way around) as few changes as possible should be required. This has
two beneficial consequences. First, the amount of effort required
to switch between simulator and real robots is reduced, and second,
the probability of new bugs being introduced during porting is minimized.
Both together improve speed and ease of developing user code for the
Kilobots.
\end{itemize}

\subsection{Implementation}

A user-controllable robot simulator consists of at least two essential
parts. The simulation core has to implement a physical model of the
robots, their interactions and their environment, while the programmer's
interface has to provide a way for users to programmatically control
the robot behaviour (and potentially parts of the simulation itself)
(Fig.~\ref{fig-simulator}).

\begin{figure*}
\centering{}\includegraphics[width=1\textwidth]{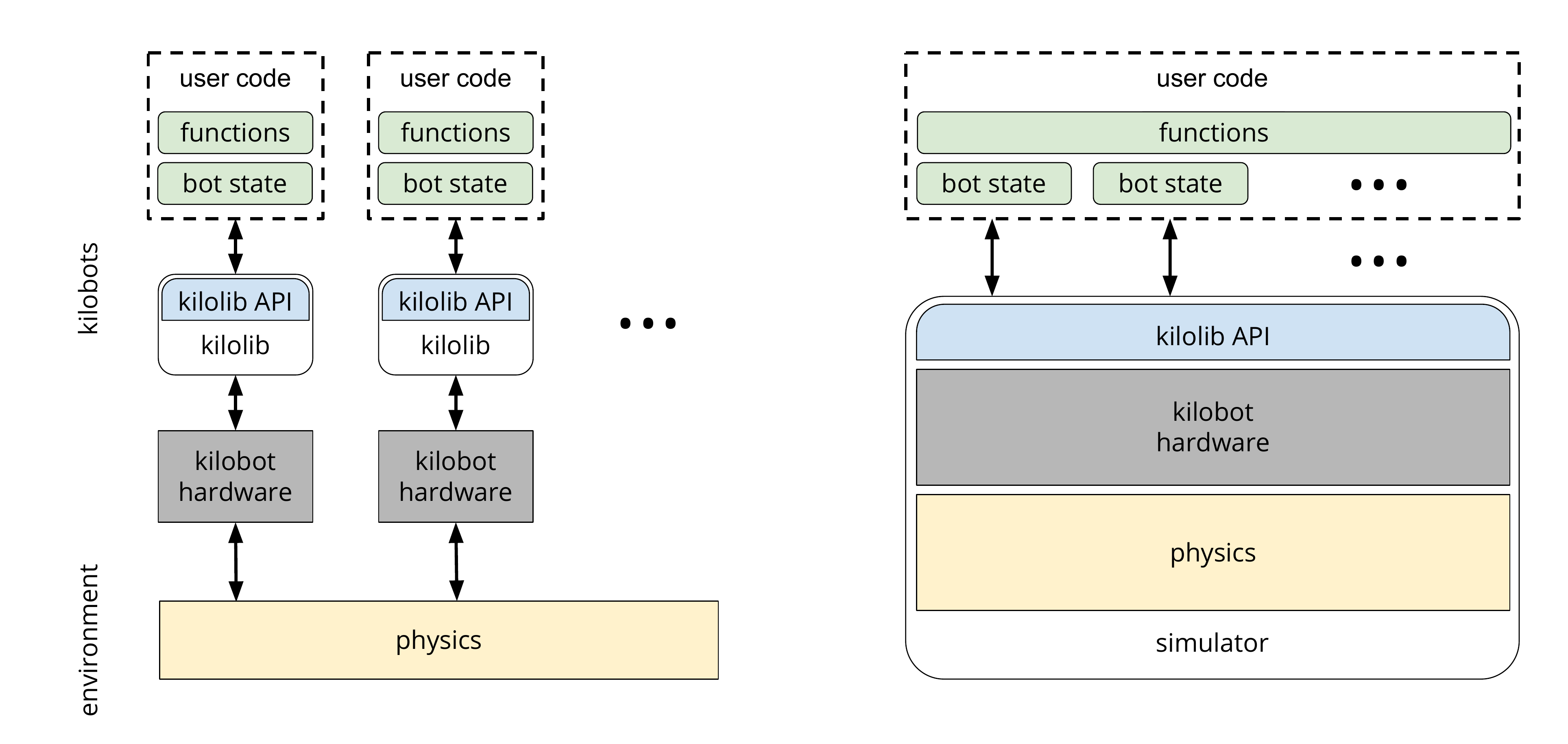}
\caption{Overview of the simulator components. The user programme in a real
robot interacts with the outside world solely through the kilolib
API (left). In the simulator (right), each robot runs its own instance
of the user programme. The simulator implements the kilolib API functions
and connects them to the simulator's representation of the physical
world.}
\label{fig-overview} 
\end{figure*}

\subsubsection{User programme}

\label{user-programme} Most common robot simulators provide users
with a scripting API in order to make development of user code faster
and easier. As stated in our requirements, however, we strive to avoid
the additional porting and debugging effort implied by this approach
and instead aimed to make user code for simulator and robots as similar
as possible while at the same time keeping the simulator as efficient
as possible.

We solved this problem by implementing the simulator Kilombo as a
nearly completely compatible drop-in replacement for kilolib, the
library that Kilobot code is compiled against. Kilombo supplies its
own version of the kilolib API that user programmes use to receive
and transmit messages and to control the robot motion. It is coupled
to the physical model, so that when a robot e.g.\ calls the API function
to turn the motors on, this robot will move forward in the physical
model. When used with Kilombo, the Kilobot C programme is natively
compiled on the simulator host, and linked with a library that implements
the physical model.

In a swarm of real Kilobots, all robot programmes run in parallel.
When using the kilolib API, the Kilobots are programmed using an event
loop. The user programme registers a loop function with the kilolib
API, and then passes control to kilolib. The loop function will then
be repeatedly called, as long as the robot is in its running state.

In our implementation, we model the parallel execution of the user
programme by sequentially calling the loop function for every robot
once per simulator time step. This greatly simplifies the simulator
design, while imposing some restrictions on the loop function, in
particular that it must return quickly without for example busy-waiting
for an external event to occur. The limitations of the chosen approach
and ways to overcome them are discussed in Sect.~\ref{portability}.

\subsubsection{Physical model}

The physical model in a robot simulator needs to incorporate those
aspects of reality which the robots interact with, those that they
can affect or observe. Given the rather basic movement and sensing
capabilities presented by Kilobots, the physical model in a Kilobot
simulator can also be approached fairly straightforwardly.

The simulator keeps track of the 2D position and the orientation of
each robot, and updates these as the simulation progresses. As explained
above, the simulation advances in time steps. In each time step, the
user programme's loop function is run once for each robot. After this,
the simulator updates the positions and orientations of the robots,
based on their movement state, which the user programme controls by
turning the two motors on or off. If both motors are on, the robot
moves forward with a constant velocity. If only one motor is on, the
robot rotates with a constant angular velocity around the rear leg
on the side opposite to the running motor. The speed and turning rate
of Kilobots depend strongly on the surface used and on the robot calibration.
In the simulator Kilombo these parameters can be configured to match
a real experiment. 

The simulator's physical model must also handle collisions between
robots. Collisions are simply resolved by displacing overlapping robots
equally along the vector joining their centres with no energy loss
to friction or plastic deformation. When real Kilobots come into contact
with each other they are often able to push each other slowly. This
pushing behaviour is captured by the simple collision dynamics in
the simulator. It also assumes perfect frictional properties between
robot legs and surface, such that each leg can move at an arbitrarily
low speed.

Communication, in the form of message passing between robots, is also
handled in the physical model. The Kilobots communicate using pulsed
infrared light. When a robot transmits a message, all other robots
within the configurable communication range receive it. This message
passing is also the mechanism by which Kilobots assess their mutual
distances -- the robot receiving the message measures the infrared
light intensity, using it to estimate the distance to the sender.

In the real world, the communication is not perfect. Sometimes messages
are lost, e.g.~if two robots close to each other transmit at the
same time. Also, the measured distance can deviate from the actual
distance. These two forms of noise are implemented in the simulator,
with a configurable probability for a message to arrive and with Gaussian
error of configurable amplitude added to each distance measurement.
An example of the effects such noise has on robot behaviour is illustrated
below in Sect.~\ref{examples}.

Kilobots can sense the ambient light level by measuring the signal
from a phototransistor. In the Kilombo simulator, ambient light sensing
is implemented using a callback mechanism to the user programme, so
that the user programme can specify a light intensity profile for
the environment. A similar mechanism is used to make user-defined
obstacles possible.

When a Kilobot turns, it rotates with one of the rear legs remaining
almost in place, as explained above. Thus the positioning of the legs
affects the turning motion of the robots. The rear legs of a Kilobot
are placed at an angle of 125\textdegree{} relative to the front leg,
measured from the centre. In the simulator the legs have the same
position by default, but can easily be moved. A configuration with
the rear legs (or more generally, the pivot points for turning) at
90\textdegree{} is relevant for a robot with two centrally placed
wheels, such as is the case for the Khepera robot from K-team. 
Allowing the pivot points to be varied within the simulator makes
it possible to explore the relative importance such robot design specificities
for the observed dynamics generated by a certain algorithm, with the
impact of the leg placement varying from minute to large, depending
on the specific algorithm.

\subsubsection{User interface}

During the simulation a graphical user interface can be displayed
that shows the swarm in bird's eye view (see Fig.~\ref{fig-simulator}).
The interface also makes it possible to interact with the simulation
at run time, e.g.~to move robots around using the mouse. The graphical
user interface is implemented using SDL (Simple DirectMedia Layer,
www.libsdl.org). During the simulation the state of the swarm can be
exported as animation frames or as numerical data. The simulation
can also be run without the graphical interface, for example for use
on a HPC cluster.

\section{Code portability}

\label{portability}

By implementing the simulator Kilombo as a drop-in-replacement for
kilolib (see Sect.~\ref{user-programme}), we made sure that user programmes
are nearly completely portable between Kilombo and the Kilobots.

Only a small number of special constructions are required for the
Kilombo simulator to handle e.g.\ global variables in the Kilobot
programme. With a few short conditionally compiled sections in the
programme, these constructions work in the physical Kilobots as well.
How the programme should be constructed is shown in the example programmes
included with the simulator and described in detail in the simulator
manual, and is also briefly explained below.

The following is a description of how a Kilobot programme should be
structured, in order to run both in a real Kilobot and in the simulator.
The issues requiring special attention are user programme variables,
timing and delays, and the data types of variables. Sect.~\ref{conversion}
shows the essence of how an existing Kilobot programme can be modified
for simulator use. Examples of the conversion of complete programmes
are given in the supplementary material as Supplement\,1. 


\lstset{language=C,
                basicstyle=\footnotesize  \ttfamily,
                aboveskip=-0.5\baselineskip,              
				keywordstyle=\color{blue}\ttfamily,                				stringstyle=\color{red}\ttfamily,
				commentstyle=\color{gray}\ttfamily,                 				
                morekeywords={uint8_t}
} 


\begin{table*}[tp]

\begin{tabular}{>{\raggedright}p{0.45\linewidth}>{\raggedright}p{0.45\linewidth}>{\raggedright}p{0.45\linewidth}}
\hline 
Original code  & Code adapted for simulator  & Comment\tabularnewline
\hline 
\begin{lstlisting}
#include <kilolib.h>
\end{lstlisting}
 & 
\begin{lstlisting}
#include <kilombo.h>
\end{lstlisting}
 & The kilombo header file assures automatic detection and compilation
for either physical Kilobots or the simulator. \tabularnewline
\midrule
\begin{lstlisting}
// Global variables
int current_motion = STOP;
int distance;
int new_message = 0;
\end{lstlisting}
 & 
\begin{lstlisting}
typedef struct {
  uint8_t current_motion; 
  uint8_t dist;
  uint8_t new_message; 
} USERDATA;
\end{lstlisting}
 & Global variables are stored in a structure. Use data types with explicit
sizes, e.g.\ \texttt{uint8\_t} and \texttt{uint16\_t}. Initialization
is done in the setup functions.\tabularnewline
\midrule
\begin{lstlisting}
// ...
\end{lstlisting}
 & 
\begin{lstlisting}
REGISTER_USERDATA(USERDATA)
// ...
\end{lstlisting}
 & This defines a pointer \texttt{mydata}, which points to the user data
structure. \tabularnewline
\midrule
\begin{lstlisting}
void loop()
{
  if (dist < TOO_CLOSE)
    {
      set_motion(FORWARD);
    }
  // ...
\end{lstlisting}
 & 
\begin{lstlisting}
void loop()
{
  if (mydata->dist < TOO_CLOSE)
    {
      set_motion(FORWARD);
    }
  // ...
\end{lstlisting}
 & Access global variables in the \texttt{USERDATA} structure through
the \texttt{mydata} pointer. \tabularnewline
\midrule
\begin{lstlisting}
// blink LED once per sec
  set_color(RGB(1, 0, 1));
  delay(500);
  set_color(RGB(0, 0, 0));
  delay(500);
}
\end{lstlisting}
 & 
\begin{lstlisting}
  if (kilo_ticks%31 < 16)
    set_color(RGB(1, 0, 1));
  else
   set_color(RGB(1, 0, 1));
}
\end{lstlisting}
 & Use \texttt{kilo\_ticks} for timing rather than \texttt{delay()}. \tabularnewline
\hline 
\end{tabular}


\caption{How to rewrite a Kilobot programme so that it can be used in the Kilombo
simulator. The modified version can also directly be compiled for
the real Kilobots as well, without requiring any further changes in
the code.}
\label{conversion} 
\end{table*}

\subsection{Programme variables}

\label{program-variables}

Kilobot C code usually makes use of static or global variables to
allow these variables to persist across repeated calls to the user-supplied
loop function. These variables demand special treatment when using
Kilombo. The simulator handles all robots through a single programme
in the simulator's memory space, so a global or static variable would
end up being common to all robots. A workaround implemented in the
simulator is to keep all global variables inside a C structure, declared
in the user programme that is registered with the runtime:

\begin{lstlisting}
typedef struct 
{
    uint8_t N_Neighbors;
    ...
} USERDATA;

REGISTER_USERDATA(USERDATA)
\end{lstlisting}
In the simulator as well as on the real robots user code accesses
the data through a pointer \texttt{mydata}, e.g.\texttt{~mydata->N\_Neighbors}.

The simulator automatically maintains an instance of this data type
for each bot that is created by the programme and ensures that the
pointer is linked to the data of the correct robot before calling
any of the user programme's functions. When the programme is compiled
for a real Kilobot, the pointer is instead linked to a single (per
robot) global variable. 

The implementation details have been hidden behind a convenience macro
\texttt{REGISTER\_USERDATA} that compiles to the respective definitions
depending on the target platform.

Note that non-static local variables (i.e.~regular variables defined
inside a function) can be used in the usual way, since these are not
required to retain their values from one function call to the next.

\subsection{Timing and delays}

\label{timing-and-delays}

The simulator calls the loop function for all robots sequentially
for every time step. This means that the loop function must return
quickly -- within the time represented by one time step. In such a
set-up it is difficult to simulate a delay in one robot while letting
the others continue to run their programmes. Also a user programme
containing polling, i.e.~looping while waiting for a condition involving
other robots to be satisfied, cannot be simulated.

The kilolib API implements a delay function, which is used to pause
programme execution for a specified amount of time. Regarding the
\texttt{delay()} function, the kilolib API documentation states the
following: ``While its easy to create short delays in the programme
execution using this function, the processor of the Kilobot cannot
perform other tasks during this delay functions (sic). In general
it is preferable to use timers to create delays.'' Taking into account
that the kilolib developers themselves indicate that this feature
is deprecated, we decided to keep the simulator design simple by just
making the delay function return immediately with no effect, and instead
rely on the timer mechanism provided by kilolib. The Kilobot API implements
a timer variable, called \texttt{kilo\_ticks}, which is incremented
at a rate of 31 times per second. Almost all timed activities can
be implemented by waiting for this variable to reach a specified value.
This design has the advantage of allowing the programme to wait for
several independent events at once, which would become difficult with
a \texttt{delay}-based programme.

In our own Kilobot programmes, we have decided to use \texttt{delay()}
only for strictly hardware-related tasks, such as spinning up the
motors. The API specifies that when a motor is turned on, it should
be run at full power for 15\,ms, after which the power should be
decreased to a calibrated value. In the simulator, the spin-up time
is not strictly necessary, so for this task the delay function is
well suited.

\subsection{Data types}

\label{data-types}

A difference between the AVR C compiler used for the Kilobots and
the native C compiler used when compiling with the simulator is the
size of data types. For example, the basic integer type is 16 bits
wide on the AVR and 32 bits wide on a standard 32 or 64 bit PC. The
larger size and thus numeric range should normally not be a problem,
unless integer overflow is used on purpose. However it may lead to
code working as intended in the simulator while overflowing on the
Kilobot. A solution is to explicitly specify the size of the types,
e.g.\ declaring variables as \texttt{uint8\_t}. This is good practice
on the AVR anyway, since it lets the programmer minimize RAM memory
usage by using the smallest possible data types.

\section{Simulation accuracy}

\label{examples}

\begin{figure}

\begin{centering}
\includegraphics[width=90mm]{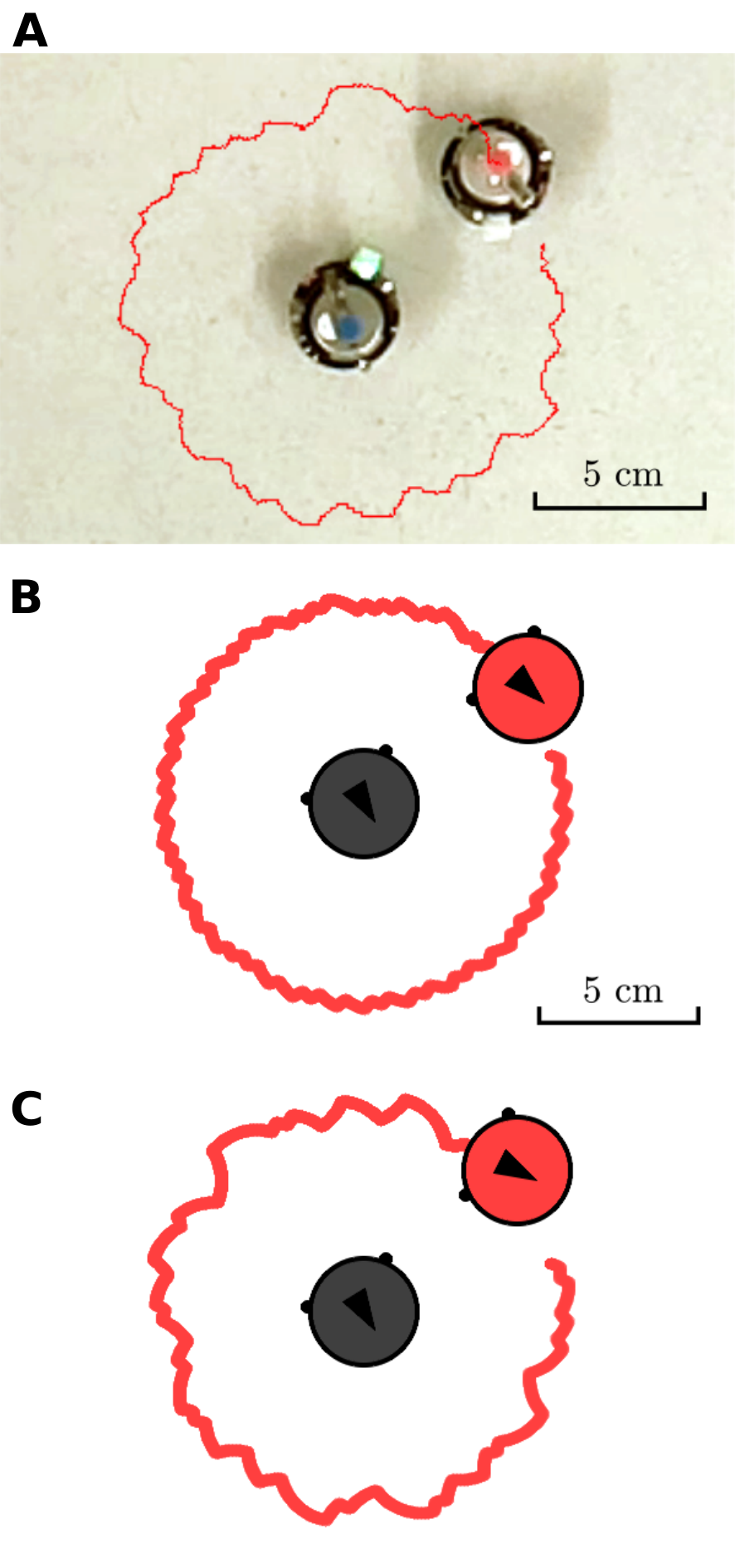} \caption{Demonstration of the ``Orbit'' programme, in which one Kilobot orbits
another one by moving while maintaining a constant distance between
them. \textbf{(A)} image of real Kilobots running the orbit programme,
with the path traced and drawn in. \textbf{(B)} Kilobots simulated
in Kilombo, running exactly the same code. \textbf{(C)} Simulation
of the same code, but with noise added in the messaging (20 \% of
the messages are lost), as well as in the distance measurement (Gaussian
noise, standard deviation of 2\,mm). The black dots on the simulated
robots represent their rear legs. See also Supplement\,2, which provides
a movie showing a direct comparison of the ``Orbit'' programme in
real Kilobots and in the simulator. }

\par\end{centering}

\label{fig-orbit1} 
\end{figure}

For a comparison between the simulated and the real Kilobots, we tested
the Kilombo simulator on the ``Orbit'' example provided with the
Kilobot documentation. Here, one robot orbits another stationary robot,
by moving while trying to keep the distance $d$ to the central robot
constant at $d_{0}=60\thinspace\text{mm}$. For orbiting a single
robot clock-wise the motion routine is simple: if $d<d_{0}$ turn
left, otherwise turn right. Fig.~\ref{fig-orbit1}A shows an image of
two Kilobots performing the orbit programme, with the path of the
moving robot traced and drawn into the still image. The path of the
orbiting robot was traced from a video recording, using simpleCV \citep{simpleCV}
to detect a red sticker attached to the centre of the charging hook.
This video is available in the supplementary material, as Supplement\,2. 

The video was recorded from above, using a Raspberry Pi with a Raspberry
Pi camera module mounted above the robots.

Fig.~\ref{fig-orbit1}B,\,C shows two different simulation runs of
the same ``Orbit'' programme that was used in Fig.~\ref{fig-orbit1}A.
Fig.~\ref{fig-orbit1}B depicts the predicted trajectory when no noise
in the communication and distance measurement is taken into account,
while Fig.~\ref{fig-orbit1}C depicts the trajectory for realistic noise
levels in the inter-robot communication. Two types of noise were introduced,
a random loss of 20\% of the messages and a Gaussian error with a
standard deviation of 2\,mm in each distance measurement. Addition
of such noise makes the simulated trajectory more similar to that
of the real robot. However, even in the noise-less simulation the
path is not smoothly circular with a constant undulation, as one in
first instance might have expected. The remaining wiggles are due
to the delay between messages and thus between distance measurements
-- the orbiting robot always moves according to the most recent distance
value obtained, which leads to small overshoots, i.e.~tangential
deviations from a circular, perfectly undulating path.

\begin{figure}
\centering{}\includegraphics[width=90mm]{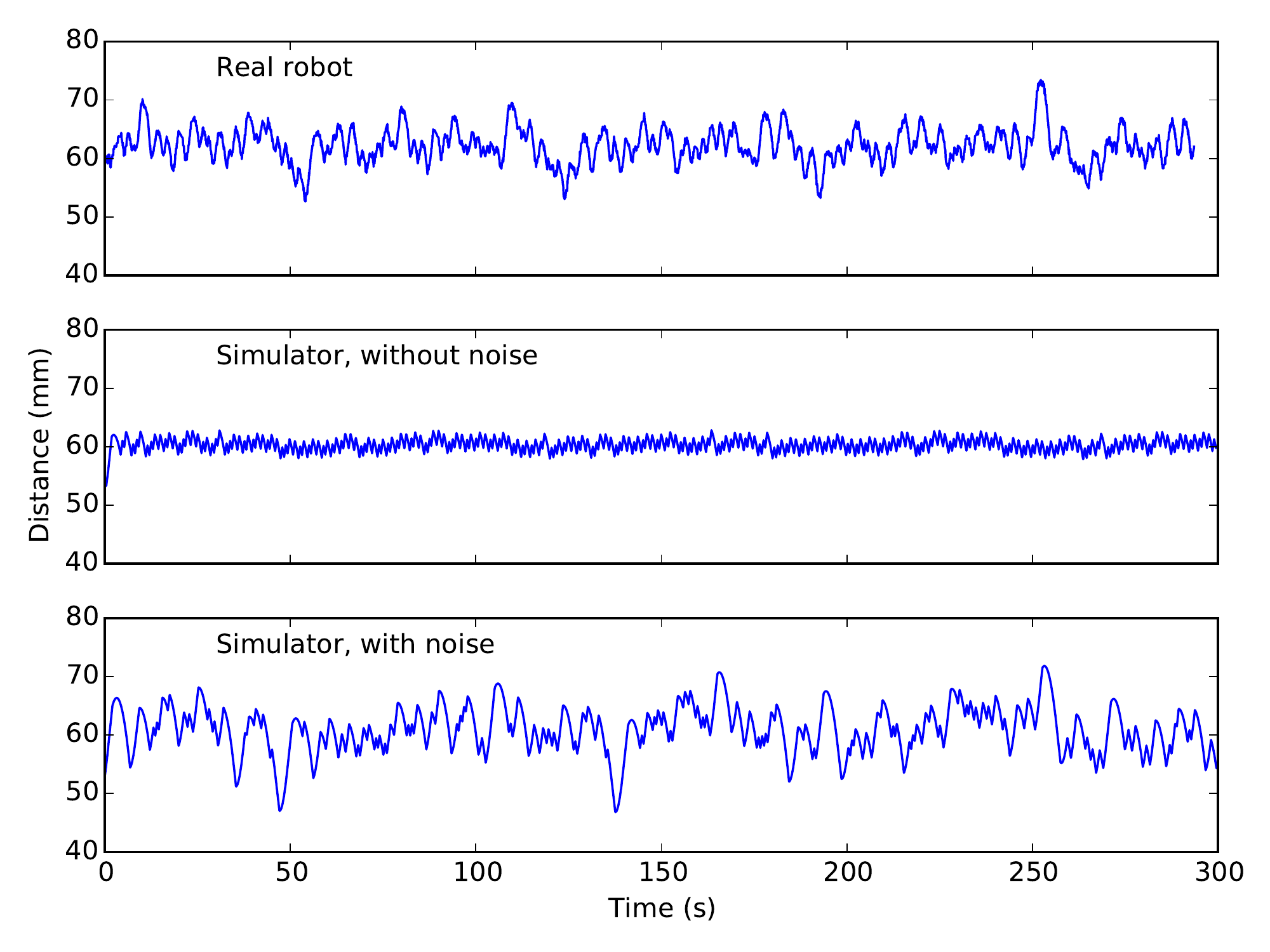}
\caption{Robot distance as a function of time for the three orbit demonstrations
shown in Fig.~\ref{fig-orbit1}. }
\label{fig-distance} 
\end{figure}

To assess the influence of noisy communication on the trajectory of
the robot in more detail, we measured the distance between the robots
as a function of time in all three Orbit experiments. These measurements
are shown in Fig.~\ref{fig-distance}, in each case for approximately
four full laps of the orbit.

\begin{figure}
\centering{}\includegraphics[width=1\textwidth]{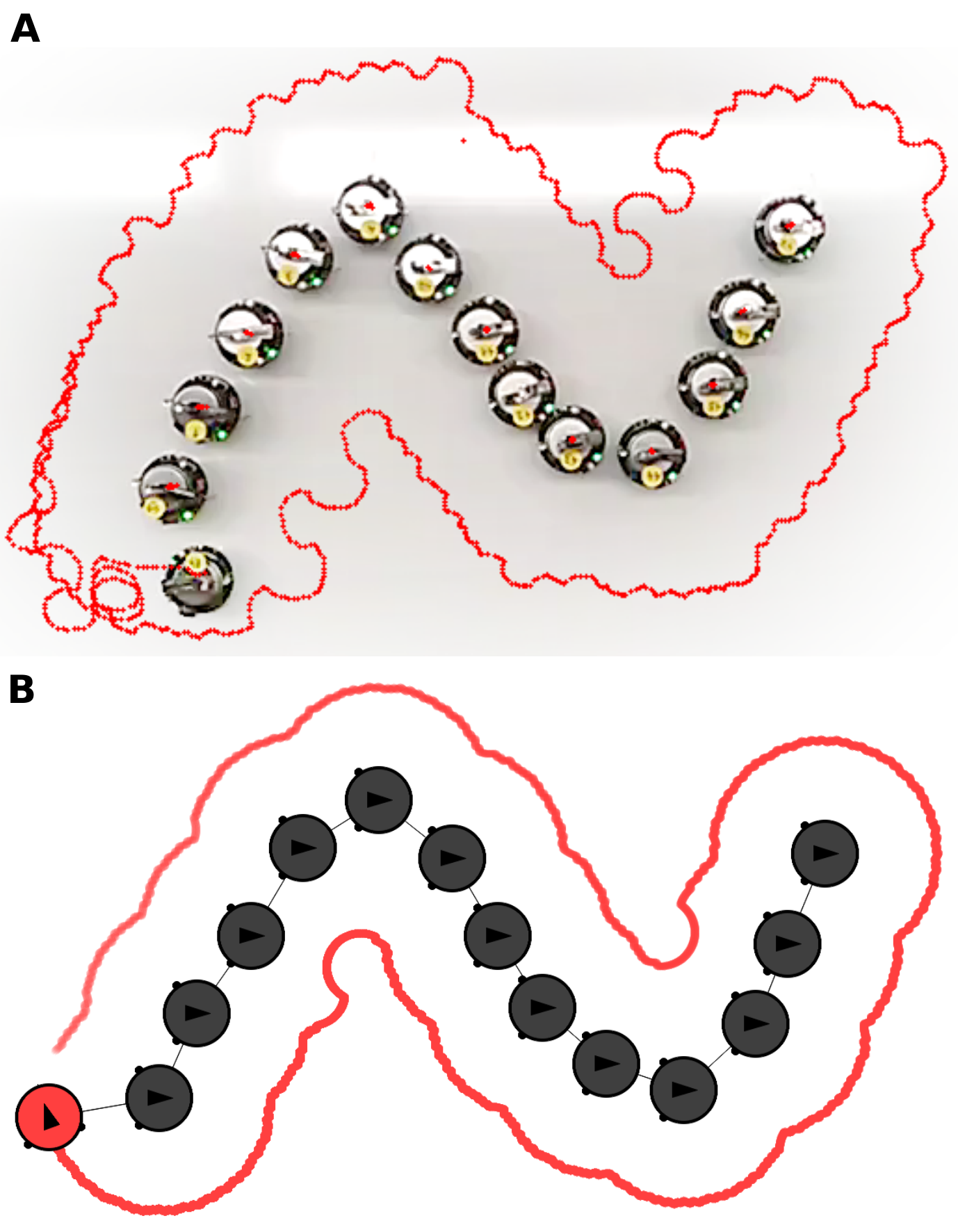}
\caption{Demonstration of the ``Edge following'' programme, in which one
Kilobot moves along the edge of a group of stationary Kilobots, by
moving while keeping a constant distance to the \emph{closest} neighbour.
\textbf{(A)} Real Kilobots running the edge following programme, with
the traced trajectory superimposed on a still image. \textbf{(B)}
Kilobots simulated in Kilombo, running the same programme.}
\label{fig-edge} 
\end{figure}

Movement while keeping a constant distance to a set of stationary
robots is an important element of the more advanced pattern-building
algorithms demonstrated with Kilobots \citep{Rubenstein2014}. If
the orbit algorithm is modified to choose the direction according
to the distance to the \emph{closest} neighbouring robot, it can be
used to move along the edge of a group of stationary robots. This
edge-following algorithm is shown in Fig.~\ref{fig-edge}, both for
real and simulated robots.

\section{Efficiency}

\label{efficiency}

When used with the Kilombo simulator, the Kilobot C programme is natively
compiled on the simulator host, and linked with the simulator library
as described in Sect.~\ref{user-programme}. Natively compiling the Kilobot
user programme makes our simulator efficient, as there is no overhead
for emulating another processor architecture or interpreting the user
programme.

\begin{figure}
\centering{}\includegraphics[width=90mm]{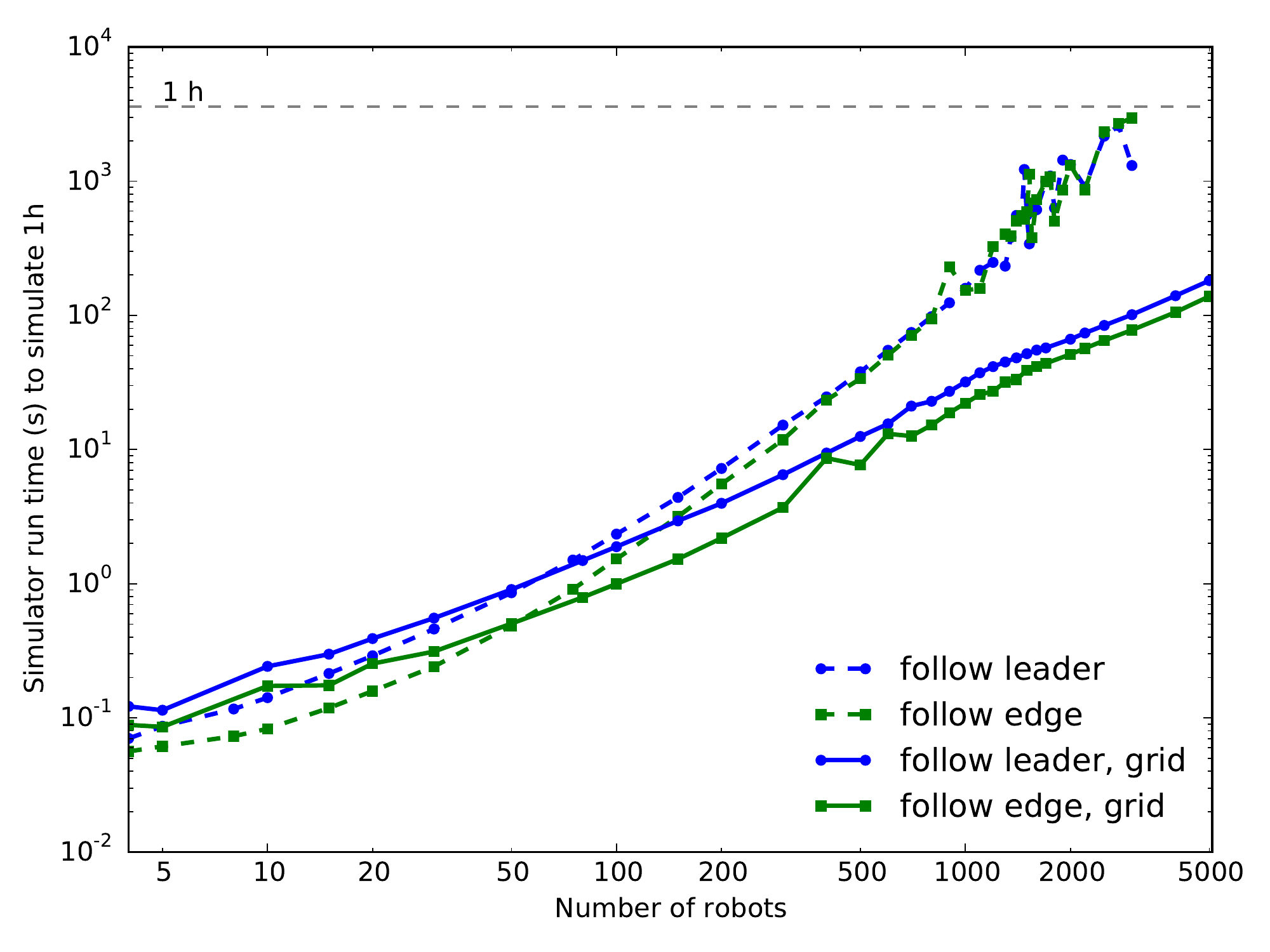}
\caption{Performance of the simulator, measured as the run time required to
simulate swarms of different sizes for 1 hour. Shown are the results
for the ``Edge following'' programme, in which a single robot moves,
and the ``Follow-the-leader'' programme, in which all robots are
moving concurrently. Dashed line indicates the 1\,h mark, i.e.~where
simulation time equals real time. }
\label{fig-performance} 
\end{figure}

To measure the Kilombo simulator performance, we recorded the run
time of simulations of two of the example programmes that are included
with the Kilombo simulator source code for different numbers of robots.
The programmes used were the \textquotedblleft Edge following\textquotedblright{}
programme (see also Fig.~\ref{fig-edge}) and the \textquotedblleft Follow-the-leader\textquotedblright{}
programme. The simulator configuration files used for the benchmark
are given in Supplement~3, where only the number of robots were changed
between the different runs. Each simulation was run for 1 hour of
simulated time without displaying the simulator GUI. The benchmarks
were made on an Intel i7-4770 system (nominally 3.40 GHz) running
Linux, with no other CPU-intensive processes running simultaneously.

In our initial implementation, the computational bottleneck turned
out to be the message passing and collision detection, specifically
the step to find which robots are within interaction range of each
other. The first implementation simply calculated the (squared) distance
between every pair of robots, and compared it to the communications
range. The performance of this implementation is shown in Fig.~\ref{fig-performance}
with dashed curves. When approximately more than 100 robots are simulated,
the neighbour-finding step dominates the computational time, causing
the runtime to scale quadratically with swarm size.

For smaller swarm sizes, the run time is also dependent on the robot
programme itself. In the edge-following simulation, only one robot
is mobile. The stationary robots execute very little code, as they
only transmit messages which the mobile robot uses to navigate. In
the follow-the-leader programme all robots are moving, each one running
a more complex programme, which makes the simulation as a whole to
run slower. For larger swarm sizes this difference then becomes dominated
by the neighbour-finding step. The irregular run-time behaviour that
can be observed around 1000 robots appears to be a cache effect. This
is the point where the data structures for all the robots no longer
fit in the 256KB L2 cache of the particular CPU in use.

To improve the performance of the neighbour-finding step, we implemented
a grid-based method for neighbour finding, the performance of which
is shown in Fig.~\ref{fig-performance} with solid curves. This grid
scheme causes the run time to scale more-or-less linearly with the
size of the swarm. The grid scheme is similar to the linked cell method
\citep{Thijssen2007} used in molecular dynamics simulations. The
scheme boils down to partitioning the 2D space in which the robots
are located into a grid, in which each grid point is a square with
side length equal to the maximal interaction range. The robots in
each grid point are stored in a list for that grid point. To find
the neighbours of a robot, one only needs to examine the robots in
the grid point containing the robot as well as robots in the eight
neighbouring grid points. With the grid method, all interacting robots
can be found in a time more-or-less linearly proportional to the number
of robots, but at the price of some additional overhead.

Fig.~\ref{fig-performance} depicts the large performance improvement
obtained when using this scheme for simulating large robot swarms.
It allows the simulator to run 100 times faster than realtime (indicated
by the dashed line) for a 1000-robot swarm. Given, however, that for
swarms with less than 50 robots the straightforward pair-wise distance
calculation is faster than the grid scheme, the simulator automatically
switches to the simpler method at small swarm sizes. Importantly,
the simulator always runs much faster than the timescale on which
the real Kilobots perform, not to mention setup and overhead time
involved in real Kilobot performances.

\section{Discussion}

\label{discussion}

\subsection{Accuracy vs.~efficiency vs.~usability}

The design of Kilombo is the result of consolidating a trade-off between
simulator complexity, accuracy, and how much the simulator must be
considered while writing the user programme. Different implementations
could have been chosen that would have led to different positions
on the trade-off spectrum.

For example, to increase accuracy the programmes for all robots could
be run truly parallel as separate threads (instead of sequentially).
It would then be possible to simulate delays in the middle of the
loop function or to explicitly account for execution time. While this
would lead to more accurate simulation, execution time, in particular
for large swarms, would suffer considerably due to the costs of synchronisation
and context switches.

An even more accurate simulation could have been achieved by building
an emulator of the Kilobots' microcontroller into the simulator. This
would additionally provide the benefit of enabling the same unaltered
binary to run on the simulator as on the Kilobots. It would however
require the implementation of a detailed model of the Kilobot hardware
down to the level of electronic components. Furthermore, emulated
code typically runs at least an order of magnitude slower than natively
compiled code.

It is important to note at this point that many aspects of the physical
Kilobot system, such as movement, communication and distance measurements,
show a high degree of stochastic variation over time and between bots.
Only gains in simulation accuracy that are at least on the same scale
as the intrinsic noise of the system will have noticeable effects
on the predicted dynamics. Furthermore, the frequency of messaging
between the robots is an order of magnitude lower than that of the
execution of the loop function. Treating the execution of the loop
function as effectively instantaneous, as is done in the Kilombo simulator,
is therefore expected to have little detrimental effect on the accuracy
of the predicted dynamics.

On the other hand, to further increase the speed of the simulations
various additional simplifications could have been implemented, such
as integrating the bot movement in between messaging events or foregoing
physics entirely and having bots move synchronously on a regular grid.
Such speed optimizations, however, would lead to significant differences
in behaviour between the simulator and the physical system.

Our implementation choices were determined by the desire to keep the
simulator as simple and efficient as possible while still being mostly
API- and behaviour-compatible to the physical Kilobots. This led to
the simple Kilombo simulator design that imposes only a few restrictions
on the user programme. We find that in practice it is relatively easy
to handle these restrictions, and we routinely use the Kilombo simulator
to develop and test our own Kilobot programmes (within the FET FoCAS
SwarmOrgan Project).

\subsection{Extension to other robots}

It would be possible to apply the Kilombo simulator also to other
robots, in particular to robots with similar simple movement and sensing
capabilities as the Kilobot. The parts of the simulator that requires
changing are the implementation of the robot's API and the connection
between the API functions and the simulator's representation of the
robots and their physical environment. We can envision a possible
extension for the AERobot educational robot \citep{Rubenstein2015},
which is based on the Kilobot, but includes more sensing possibilities
while not being directly aimed at swarm operation.

\section{Conclusions}

We have developed a C based simulator, Kilombo, that allows those
working with Kilobots to greatly speed up testing and debugging Kilobot
code. The simulator compiles the same code that runs on the physical
Kilobots, thereby removing the slow and error-prone step of converting
code to a different platform. Simulated Kilobots show good agreement
with physical Kilobots running the same programme, given some constraints
on movement patterns. The simulator makes high-throughput pre-screening
possible of potential algorithms that could lead to desired emergent
behaviour. We argue that this strategy, here specifically developed
for Kilobots, is of general importance for effective robot swarm research.
The code is freely available \citep{sourcecode} under the MIT license.

\begin{acknowledgements}

\label{acknowledgments} Valuable comments on the manuscript by Stéphane
Doncieux and Yaochu Yin are gratefully acknowledged. The Kilombo Kilobot
simulator has been developed within the Swarm-Organ project, funded
by the European Union Seventh Framework Programme (FP7/2007-2013)
under grant agreement 601062. J.S., N.C. and I.S. acknowledge support
of the Spanish Ministry of Economy and Competitiveness, `Centro de
Excelencia Severo Ochoa 2013-2017', SEV-2012-0208. AFMM was supported
by the UK Biological and Biotechnology Research Council (BBSRC) via
grant BB/J004553/1 to the John Innes Centre.
\end{acknowledgements}


\bibliographystyle{springer/spbasic}
\bibliography{simulator}

\end{document}